\documentclass[10pt,twocolumn,letterpaper]{article}

\usepackage{iccv}
\usepackage{times}
\usepackage{epsfig}
\usepackage{graphicx}
\usepackage{amsmath}
\usepackage{amssymb}
\usepackage{algpseudocode}
\usepackage{booktabs}
\usepackage{cuted}
\usepackage{capt-of}

\usepackage{array}
\newcolumntype{P}[1]{>{\centering\arraybackslash}p{#1}}
\usepackage{xcolor,colortbl}
\definecolor{Gray}{gray}{0.9}
\usepackage{cite}
\usepackage{pifont}
\usepackage{multirow}

\usepackage[pagebackref=true,breaklinks=true,letterpaper=true,colorlinks,bookmarks=false]{hyperref}

\iccvfinalcopy 

\def\iccvPaperID{****} 

\ificcvfinal\fi

\begin{document}

\title{Boundary-aware Camouflaged Object Detection via Deformable Point Sampling}

\author{Minhyeok Lee$^{1}$\quad Suhwan Cho$^{1}$\quad Chaewon Park$^1$\quad Dogyoon Lee$^{1}$\quad Jungho Lee$^{1}$\quad Sangyoun Lee$^{1,2}$\vspace{0.5cm}\\
	$^1$~~Yonsei University\\
	$^2$~~Korea Institute of Science and Technology (KIST)}
\maketitle
\ificcvfinal\thispagestyle{empty}\fi

\begin{abstract}	
	The camouflaged object detection (COD) task aims to identify and segment objects that blend into the background due to their similar color or texture. Despite the inherent difficulties of the task, COD has gained considerable attention in several fields, such as medicine, life-saving, and anti-military fields. In this paper, we propose a novel solution called the Deformable Point Sampling network (DPS-Net) to address the challenges associated with COD. The proposed DPS-Net utilizes a Deformable Point Sampling transformer (DPS transformer) that can effectively capture sparse local boundary information of significant object boundaries in COD using a deformable point sampling method. Moreover, the DPS transformer demonstrates robust COD performance by extracting contextual features for target object localization through integrating rough global positional information of objects with boundary local information. We evaluate our method on three prominent datasets and achieve state-of-the-art performance. Our results demonstrate the effectiveness of the proposed method through comparative experiments.
\end{abstract}

\section{Introduction}
\label{sec:intro}
Camouflaged object detection (COD) is an image segmentation task that aims to identify objects that are challenging for human detection due to their similarity in appearance to the background. COD has recently been used in various applications, such as polyp segmentation in the medical field, lifesaving in extreme environments, and antimilitary camouflage. COD is similar to salient object detection (SOD) in that there is no prior information about the target object; however, it is a more challenging task because the target is camouflaged. Therefore, the detection of subtle differences in color and pattern of camouflaged objects is crucial for successful COD.

\begin{figure}[t]
	\setlength{\belowcaptionskip}{-24pt}
	\begin{center}
		\includegraphics[width=0.9\linewidth]{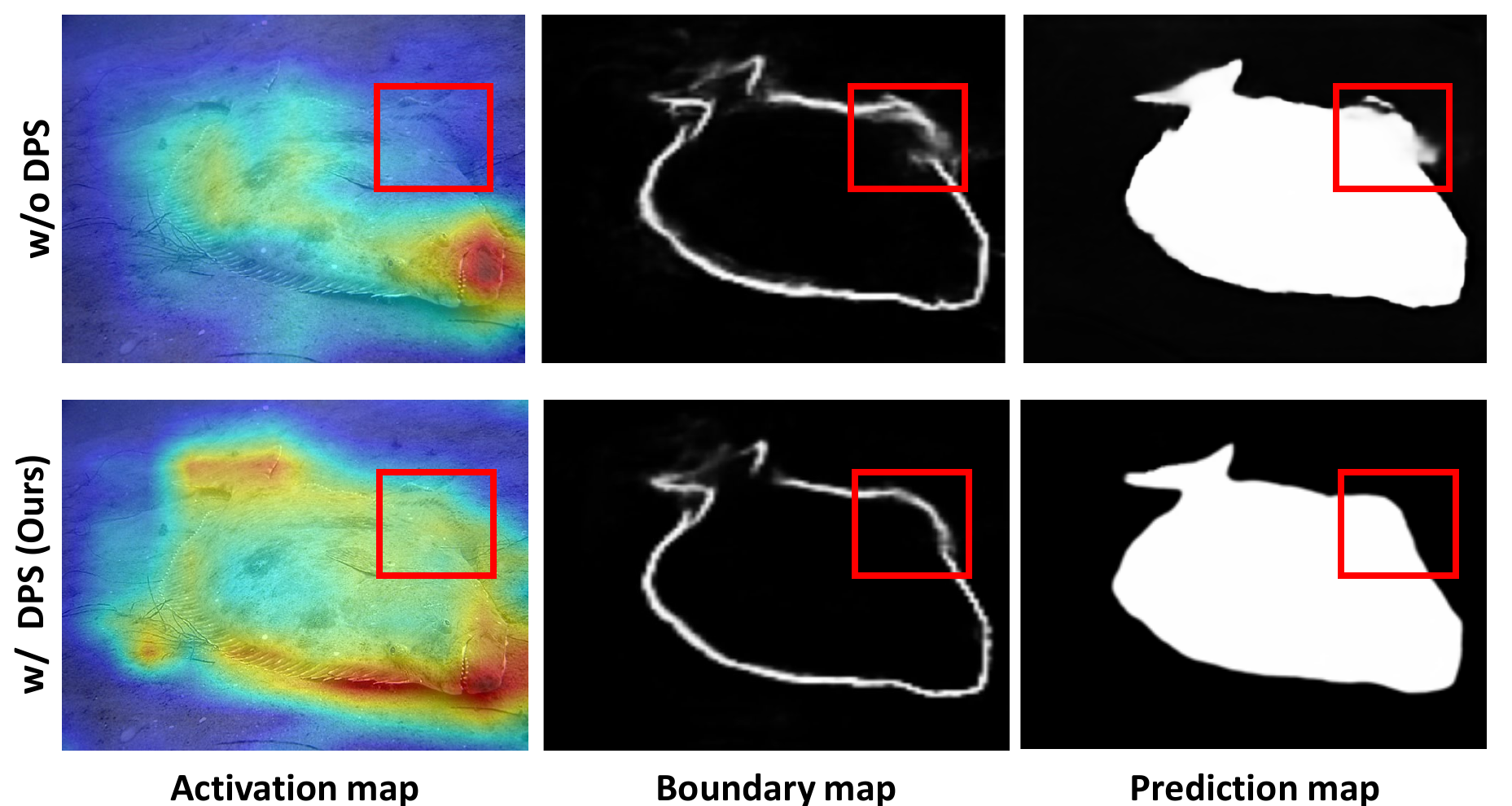}
		\caption{Effectiveness of the proposed DPS transformer. The visualization shows the activation maps of the features learned by the naive encoder without DPS transformer (Top), and the features learned with the DPS transformer (Bottom).}
		\vspace{-0.5cm}
		\label{fig:intro}
	\end{center}
\end{figure}

Various studies~\cite{fan2020camouflaged, ren2021deep, zhu2021inferring, sun2021context, sun2022boundary, jia2022segment, zhu2022can, ji2022fast} propose deep learning-based methods to address the challenges of COD. Some approaches~\cite{ren2021deep, zhu2021inferring} extract global texture features of the camouflaged object to learn its information separated from the background. However, since the texture of most camouflaged objects is similar to the background, distinguishing subtle differences in boundary local information remains a challenging task. Jia~\textit{et al.}~\cite{jia2022segment} use a multistage prediction method that predicts the camouflaged object after magnifying the image based on its pre-predicted position. However, this method suffers from the drawback of having a large inference time because the same model is used repeatedly. Recent studies~\cite{sun2022boundary, zhu2022can, ji2022fast} improve model performance by introducing an additional module that reconstructs boundaries from encoder features. Typically, these approaches attach boundary prediction branches to the encoder to provide additional supervision signals for object edges. These techniques can encourage the encoder to focus more on the object boundary regions during the training phase. However, due to the limitations of the encoder's feature extraction capabilities, low-quality predictive masks are generated in the boundary regions when the camouflaged object and the background are very similar, as shown in the activation map in Figure~\ref{fig:intro} (a).

To solve these problems, we propose the Deformable Point Sampling network (DPS-Net). Our model employs a Deformable Point Sampling transformer (DPS transformer) to directly capture sparse supervision signals for object boundary areas from the encoder features. The DPS transformer consists of a local extractor, a global extractor, and an aggregator. Firstly, the local extractor divides the image into patches and extracts local information from deformable sampling points using offset encoders. This feature sampling method encourages the extraction of more information in meaningful areas of the disguised object boundaries through predicted offsets. Thus, the local extractor can learn detailed local information of camouflaged objects and distinctive features from the background while increasing the computational efficiency of the model by sampling only useful information. Secondly, the global extractor creates soft object region masks for the objects and the background from the input features. Global features are extracted from each feature map area by applying global average pooling to the generated mask region. This process enables effective extraction of global features of camouflaged objects and specifies their approximate location. Finally, the aggregator aggregates the generated local and global features. Although local information about boundaries is critical in COD, focusing only on local information can overlook important global information for roughly estimating the location of the target object. Therefore, we design an aggregator to simultaneously utilize global and local information. Figure~\ref{fig:intro} (b) demonstrates that the proposed DPS transformer effectively complements the feature extraction capability of the encoder and integrates global positional information and local boundary information. As a result, the DPS transformer identifies the overall tendency of the image by considering the global context and local details of the image simultaneously. Moreover, it helps in understanding boundaries by capturing the fine structure of the object and consequently allows the model to generate an accurate prediction map.

We tested our method on three popular datasets: CHAMELEON~\cite{skurowski2018animal}, CAMO~\cite{le2019anabranch}, and COD10K~\cite{fan2020camouflaged}. These datasets contain various challenging scenarios, and the proposed model achieves state-of-the-art performance on all three datasets. In addition, we performed various ablation studies to prove the effectiveness of the model and show that robust COD is possible in challenging scenes.

Our main contributions can be summarized as follows:
\begin{itemize}
	\item We propose a novel DPS network with a DPS transformer to overcome the limitations of the previous COD model. The DPS transformer directly capture sparse supervision signals for object boundary areas from the encoder features.
	
	\item The proposed model effectively combines local boundary information using deformable sampling points and global information of the object to generate accurate boundary maps and prediction masks.
	
	\item The proposed network achieves state-of-the-art performance on the CHAMELEON~\cite{skurowski2018animal}, CAMO~\cite{le2019anabranch}, and COD10K~\cite{fan2020camouflaged} datasets. Additionally, we demonstrate the effectiveness of the proposed method through various ablation studies.
\end{itemize}

\section{Related Work}
\noindent
\textbf{Camouflaged Object Detection} Traditional COD methods~\cite{kavitha2011efficient, bhajantri2006camouflage, huerta2007improving, feng2013camouflage} use hand-crafted features such as color, boundary, texture, convex, and brightness to distinguish camouflaged objects from their backgrounds. However, these hand-based methods have poor detection results when the target object has colors and textures that are very similar to those of the background.

To solve this problem, deep learning-based COD methods~\cite{fan2020camouflaged, ji2022fast, zhu2022can, ren2021deep, zhu2021inferring, zhong2022detecting, sun2021context, sun2022boundary, jia2022segment} have been recently proposed. For example, SINet~\cite{fan2020camouflaged} approaches COD as an SOD problem and applies sophisticated SOD technology to the network. Some studies~\cite{ren2021deep, zhu2021inferring} extract texture information to separate the target object from the background. However, these methods often fail when the target object shares the same texture features with the background. Jia~\textit{et al.}~\cite{jia2022segment} uses a network to specify the approximate location of the target object and repeats the process of magnifying and cropping the image based on this approximate location. However, their iterative multi-stage method has the disadvantage of significantly increasing the model's inference time. Zhong~\textit{et al.}~\cite{zhong2022detecting} proposes a frequency-enhancement module to focus on feature points in the frequency domain that are distinct from the background. Furthermore, several recent papers~\cite{zhai2021mutual, sun2022boundary, zhu2022can, ji2022fast} improve the model's performance by reconstructing the boundaries of the target object from encoder features and integrating them. However, because boundary information is directly extracted from the encoder, there is a problem that the quality of the reconstructed boundary is dependent on the quality of the encoder feature.

\begin{figure*}[t]
	\setlength{\belowcaptionskip}{-24pt}
	\begin{center}
		\includegraphics[width=0.9\linewidth]{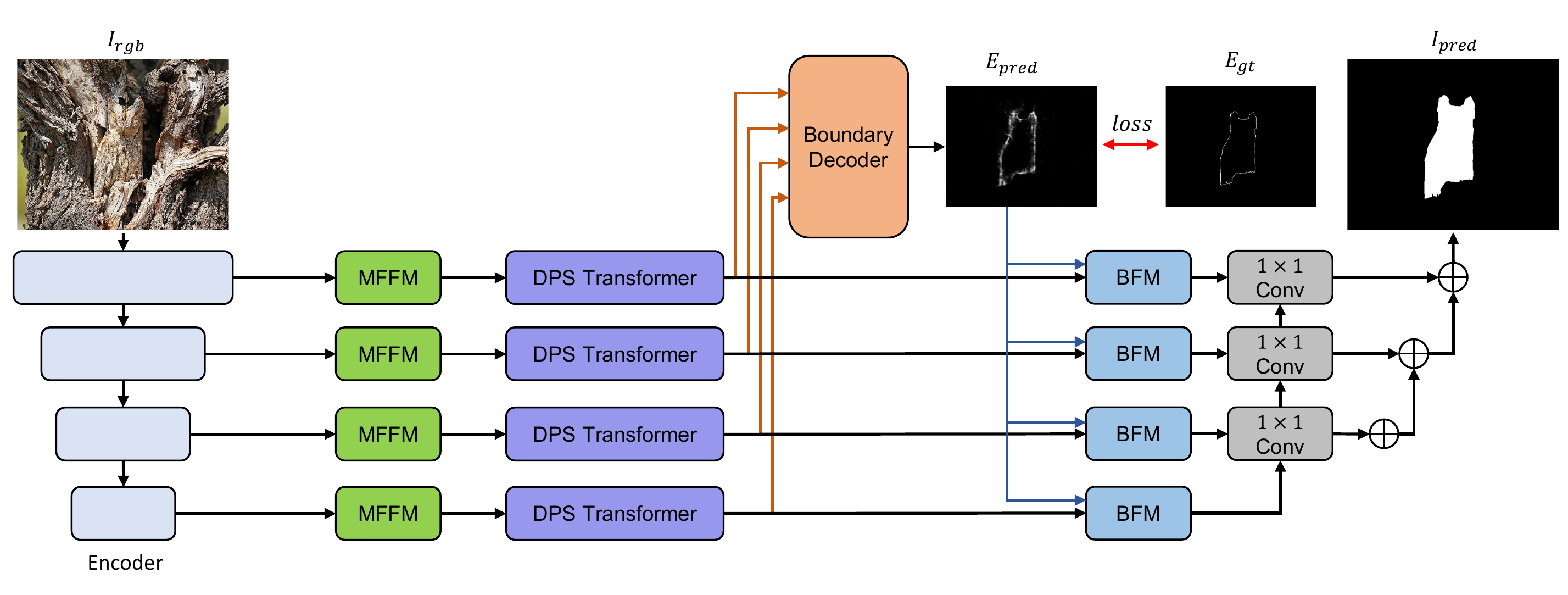}
		\caption{Overall architecture of the DPS-Net. The MFFM extracts and integrates multi-scale features from encoders. The DPS transformer extracts detailed local boundary information and coarse global information of the target object from the encoder features and aggregates them. The boundary decoder extracts the boundary of camouflaged objects. The BFM effectively fuses features generated from DPS transformers and boundary maps.}
		\vspace{-0.5cm}
		\label{fig:main}
	\end{center}
\end{figure*}

\section{Proposed Approach}
\subsection{Overall Architecture}
Figure~\ref{fig:main} shows the overall architecture of DPS-Net. The proposed model comprises an RGB encoder, a boundary decoder for boundary reconstruction, and a feature pyramid network-based~\cite{lin2017feature} decoder for generating a final segmentation map. As a first step, the proposed model extracts and integrates multi-scale features from encoder blocks using multi-scale feature fusion modules (MFFMs). Next, the proposed DPS transformer block extracts global and local features of target objects and aggregates those features. The boundary decoder predicts the boundary map $E_{pred}$, and $E_{pred}$ is merged with the aggregated features through the boundary fusion modules (BFMs). Finally, the model generates a final prediction mask $I_{pred}$ through a decoder.

\begin{figure}[t]
	\setlength{\belowcaptionskip}{-24pt}
	\begin{center}
		\includegraphics[width=0.9\linewidth]{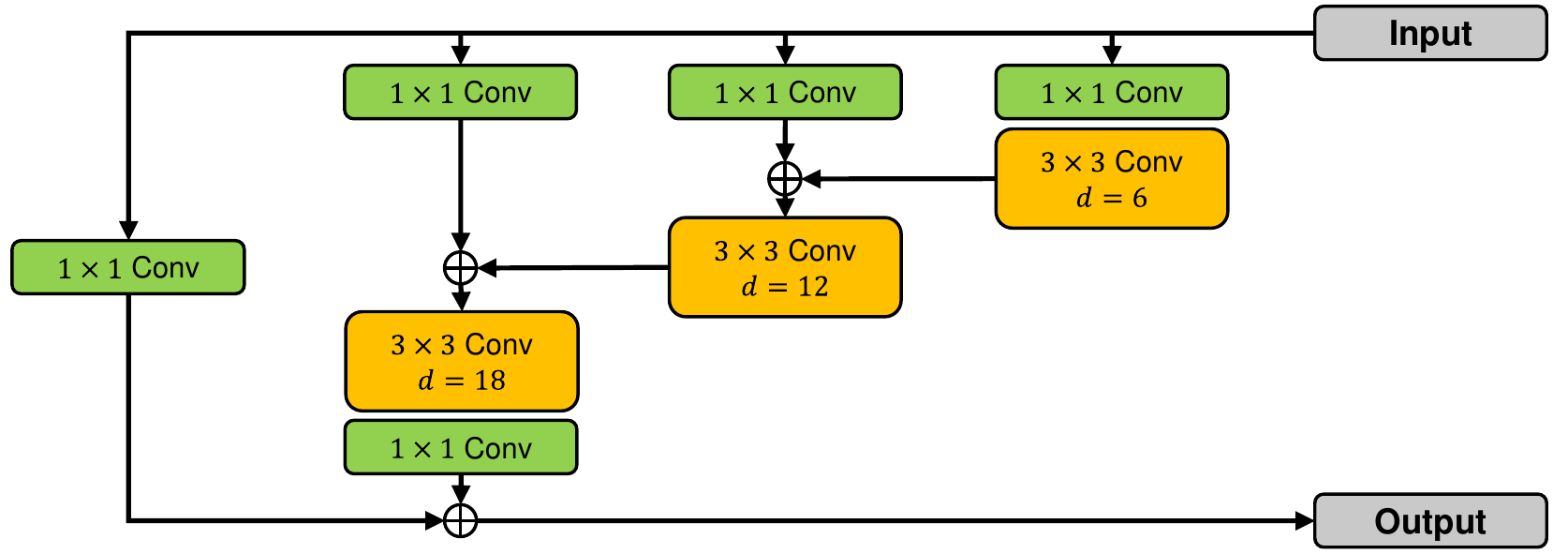}
		\caption{Structure of the proposed MFFM. The MFFM effectively integrates multi-scale features by sequentially applying dilated convolutions of different ratios.}
		\vspace{-0.5cm}
		\label{fig:MFFM}
	\end{center}
\end{figure}

\subsection{Multi-Scale Feature Fusion Module}
We use MFFM for effectively extracting and integrating multi-scale features from encoders. Inspired by atrous spatial pyramid pooling~\cite{chen2017deeplab}, MFFM comprises dilated convolutional layers with different ratios as shown in Figure~\ref{fig:MFFM}. In particular, we apply $3 \times 3$ dilated convolutions with ratios of 6, 12, and 18, respectively. Further, encoder features are sequentially integrated from the small to the large ratio. This structure effectively increases the receptive field with a small number of parameters and delivers rich multi-scale contextual information to the GLA transformers, which will be described later.

\subsection{Deformable Point Sampling Transformer}
\label{sec:tf}
\begin{figure*}[t]
	\setlength{\belowcaptionskip}{-24pt}
	\begin{center}
		\includegraphics[width=\linewidth]{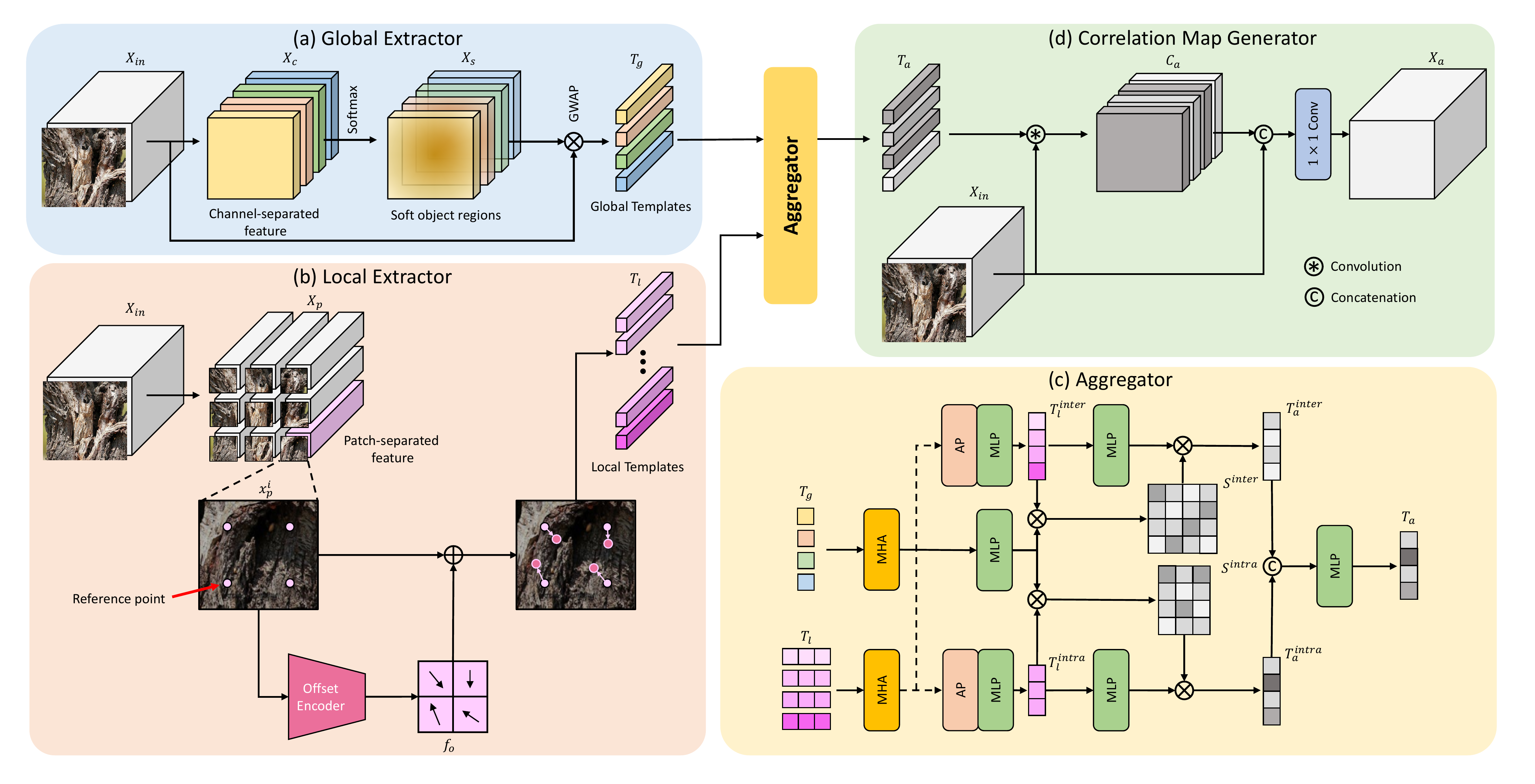}
		\caption{Structure of the DPS transformer, composed primarily of four subparts. (a) The global extractor generates global templates of input features. (b) The local extractor separates the input features into patches and extracts the local templates from each patch. (c) The aggregator aggregates the extracted global and local templates. (d) The correlation map generator generates correlation maps from the aggregated features.}
		\vspace{-0.5cm}
		\label{fig:GLT}
	\end{center}
\end{figure*}

The DPS transformer aims to effectively extract useful boundary local context features for mask prediction of camouflaged objects and fuse them with global object features. As shown in Figure~\ref{fig:GLT}, the proposed DPS transformer is mainly composed of a global extractor, a local extractor, an aggregator, and a correlation map generator.

\noindent
\textbf{Global Extractor.} The global extractor generates global templates $\mathbf{T_g}$ from the input feature $\mathbf{X_{in}} \in \mathbb{R} ^ {C \times H \times W}$, where $C$, $H$, and $W$ indicate the channel, height, and width of the input feature, respectively. As a first step, the global extractor separates each channel of the input feature to generate a channel-separated feature, as shown in Figure~\ref{fig:GLT} (a). In particular, the $i$-th channel of $\mathbf{X_{in}}$ is defined as $\mathbf{X^{i}_{c}} \in \mathbb{R} ^ {H \times W}$. Next, a channel-wise softmax operation is applied to create soft object regions~\cite{yuan2020object} $\mathbf{X^{i}_{s}} \in \left[0, 1\right] ^ {H \times W}$. According to~\cite{yuan2020object}, as each channel in $\mathbf{X_{in}}$ is generated from the convolutional kernels of the trained encoder, $\mathbf{X_{s}}$ contains approximate areas for background or foreground objects. Then, to generate global features, a global weighted average pooling (GWAP)~\cite{qiu2018global} operation is performed with $\mathbf{X_{in}}$, treating $\mathbf{X^{i}_{s}}$ as a weighted mask. In other words, the global template $\mathbf{t _ { g } ^ { i }} \in \mathbb{R} ^ {C}$ generated by $\mathbf{X^{i}_{s}}$ is expressed as follows:

\begin{equation}
	\mathbf{t _ { g } ^ { i }} =GAP \left ( \mathbf{X _ { in }} \cdot \mathbf{X _ { s } ^ { i }} \right ),
\end{equation}

\noindent
where $i = 1, 2, ..., C$, and $GAP \left ( . \right )$ is a global average pooling operator. Finally, the global extractor creates a global template block $\mathbf{T_{g}}$, which is a set of generated global templates $\mathbf{t _ { g } ^ { i }}$. This method observes the image as a whole and extracts representative features of the scene based on the semantic context. Therefore, $\mathbf{T_{g}}$ contains global information and thus includes prior knowledge to distinguish objects or backgrounds. Because the number of channels in $\mathbf{X _ { s }}$ is $C$, the size of $\mathbf{T_{g}}$ is $C \times C$.

\noindent
\textbf{Local Extractor.} The local extractor separates the input features into patches and extracts the local features for each patch. As shown in Figure~\ref{fig:GLT} (b), the local extractor splits the input feature $\mathbf{X_{in}}$ into $N_{p} \times N_{p}$ patches. Therefore, the size of the $i$-th patch $\mathbf{x_{p}^i}$ is $C \times \frac{ H } { N_{p} } \times \frac{ W } { N_{p} }$. However, like many vision transformer methods~\cite{dosovitskiy2020image, zheng2021rethinking, strudel2021segmenter, xie2021segformer}, extracting features from every pixel in each patch and applying a transformer is computationally expensive and makes model convergence difficult. To solve this problem, we propose a local feature extraction method inspired by deformable attention~\cite{zhu2020deformable}. First, $N_r \times N_r$ reference points are initialized uniformly on $\mathbf{x_{p}^i}$ as shown in the Figure~\ref{fig:GLT} (b). Next, $\mathbf{x_{p}^i}$ is feed into a small offset encoder $\theta_{off}$ to create an offset field. $\theta_{off}$ includes two convolutional layers and one GeLU~\cite{hendrycks2016gaussian} layer between them, and the tangent hyperbolic function (tanh) is used as the output activation function of $\theta_{off}$. In addition, a predefined factor $s$ is applied to prevent drastic movement of reference points and stabilize learning. In other words, the offset field $\mathbf{f_{o}} \in \left( -s, +s \right) ^ {2 \times N_r \times N_r}$ is expressed as follows:

\begin{equation}
	\mathbf{f_{o}} =s \times tanh \left ( \theta _ { off } \left ( \mathbf{x _ { p } ^ { i }} \right ) \right ),
\end{equation}

\begin{figure*}[t]
	\setlength{\belowcaptionskip}{-24pt}
	\begin{center}
		\includegraphics[width=\linewidth]{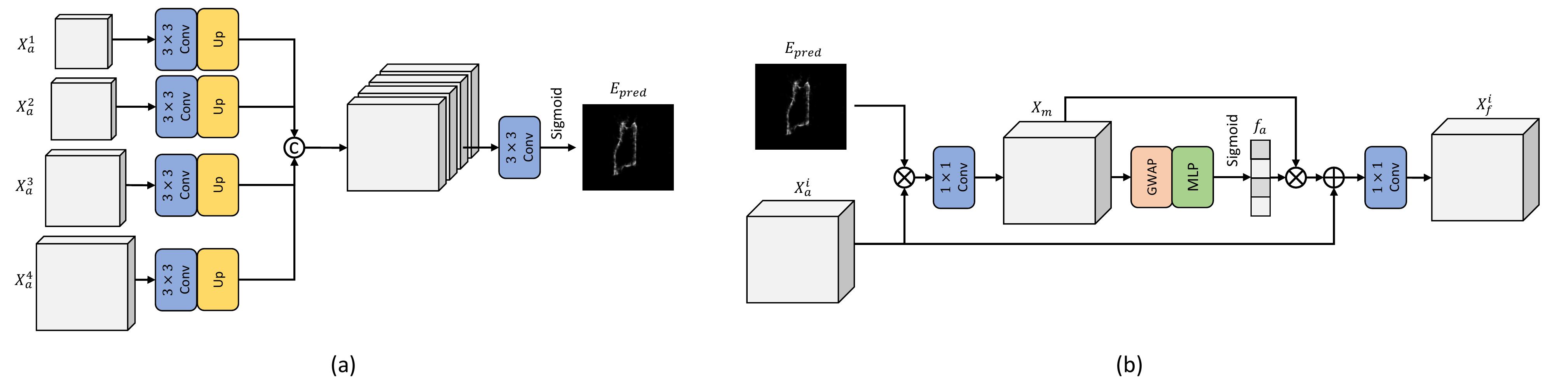}
		\caption{Structures of the proposed boundary decoder (a) and boundary fusion module (BFM) (b). The boundary decoder aims to extract the boundary of camouflaged objects. BFM integrates the features from DPS transformers and the boundary map.}
		\vspace{-0.5cm}
		\label{fig:boundary}
	\end{center}
\end{figure*}

\noindent
where $\mathbf{f_{o}}$ represents the relative amount of change in the x- and y-axis directions of each reference point, where $\mathbf{x_{p}^i}$ is regarded to have a size of $1 \times 1$. Finally, each reference point is moved according to $\mathbf{f_{o}}$ and the features of size $C  \times 1 \times 1$ corresponding to that pixel are sampled. However, as exactly locating the moved reference points on a specific pixel on $\mathbf{x_{p}^i}$ is impossible, we follow~\cite{zhu2020deformable} to sample the feature by applying bilinear interpolation to 4 adjacent pixels. Local templates generated from $\mathbf{x_{p}^i}$ are defined as $\mathbf{t_{l}^i}$. Further, the local template $\mathbf{T_{l}}$ generated from the entire patches is a set of $\mathbf{t_{l}^1}, \mathbf{t_{l}^2}, ..., \mathbf{t_{l}}^{\left( N _ { p } \times N _ { p } \right)}$. Therefore, as $N_r \times N_r$ local templates are sampled for one patch, the size of $\mathbf{T_{l}}$ sampled from the entire patch is $C \times \left ( N _ { p } \times N _ { p } \right ) \times \left ( N _ { r } \times N _ { r } \right )$. Consequently, $\mathbf{T_{l}}$ contains local features extracted from each patch. Moreover, instead of storing information for every pixel, only key features are stored according to deformable attention for computational efficiency.

\noindent
\textbf{Aggregator.} The aggregator aims to generate useful features for camouflaged object mask reconstruction by effectively aggregating the extracted global templates $\mathbf{T_{g}}$ and local templates $\mathbf{T_{l}}$. Therefore, for a given module, considering the relations between global and local templates, between patches in local extractors, and between local templates within patches is important. Therefore, we design aggregator inspired by CurveNet~\cite{xiang2021walk} in the 3D point cloud classification task. As shown in Figure~\ref{fig:GLT} (c), first key, query, and value-based multi-head attention~\cite{wang2018non, fu2019dual, zhang2019self} are applied to enhance the correlation between $\mathbf{T_{g}}$ templates and the correlation between $\mathbf{T_{l}}$ templates. Next, an attentive pooling (AP)~\cite{hu2020randla} operation is applied along each axis of $\mathbf{T_{l}}$ to generate an inter-patch local feature $ \mathbf{T_{l}^{inter}} \in \mathbb{R} ^ {C \times \left ( N _ { r } \times N _ { r } \right )}$ and an intra-patch local feature  $ \mathbf{T_{l}^{intra}} \in \mathbb{R} ^ {C \times \left ( N _ { p } \times N _ { p } \right )}$. Further, $\mathbf{T_{g}}$, $\mathbf{T_{l}^{inter}}$, and $\mathbf{T_{l}^{intra}}$ are fed to individual multi-layer perceptrons (MLPs). In addition, as shown in Figure~\ref{fig:GLT} (c), matrix multiplication and softmax operation are applied among $\mathbf{T_{g}}$, $\mathbf{T_{l}^{inter}}$, and $\mathbf{T_{l}^{intra}}$ to generate two correlation score maps $\mathbf{S^{inter}} \in {\left(0, 1 \right)} ^ {C \times \left ( N _ { r } \times N _ { r } \right )}$ and $\mathbf{S^{intra}} \in {\left(0, 1 \right)} ^ {C \times \left ( N _ { p } \times N _ { p } \right )}$.  In another branch, $\mathbf{T_{l}^{inter}}$ and $\mathbf{T_{l}^{intra}}$ are further transformed with two extra MLPs, which are then fused with the correlation score maps by matrix multiplication separately. With the above process, two types of aggregated features--$\mathbf{T_{a}^{inter}} \in \mathbb{R} ^ {C \times C}$ and $\mathbf{T_{a}^{intra}} \in \mathbb{R} ^ {C \times C}$-- are created; finally, $\mathbf{T_{a}} \in \mathbb{R} ^ {C \times C}$ are fused by the MLP layer.

\noindent
\textbf{Correlation Map Generator.} The correlation map generator generates correlation features from $\mathbf{T_{a}}$. Each template of $\mathbf{T_{a}}$ is treated as a $1 \times 1$ convolution kernel and convolution is performed with $\mathbf{X_{in}}$. Because $\mathbf{T_{a}}$ contains a total of $C$ templates, the size of the generated correlation feature $\mathbf{C_{a}}$ is $C \times H \times W$. Finally, $\mathbf{C_{a}}$ and $\mathbf{X_{in}}$ are concatenated and the final output feature $\mathbf{X_{a}} \in \mathbb{R} ^ {C \times H \times W}$ is generated through $1 \times 1$ convolution.

\subsection{Boundary Decoder}
Figure~\ref{fig:boundary} (a) shows the structure of the proposed boundary decoder. The boundary decoder aims to effectively extract the boundary of camouflaged objects from aggregated multi-scale features $\mathbf{X_{a}^{1}}$, $\mathbf{X_{a}^{2}}$, $\mathbf{X_{a}^{3}}$, and $\mathbf{X_{a}^{4}}$ from the DPS transformers. The boundary decoder comprises a $3 \times 3$ convolution layer and upsampling layers, integrating the multiscale features. Finally, all the features are concatenated and passed to the convolutional and sigmoid layers to create a single-channel boundary map $\mathbf{E_{pred}}$.

\subsection{Boundary Fusion Module}
We propose the BFM to integrate the features from the DPS transformers and boundary map with different levels of feature representation using boundary information as a guidance. As shown in Figure~\ref{fig:boundary} (b), the input of the proposed BFM comprises the boundary map $\mathbf{E_{pred}}$ generated by the boundary decoder and the feature $\mathbf{X_{a}^{i}}$ generated by the $i$-th DPS transformer, both of which are resized to the same resolution. Next, $\mathbf{E_{pred}}$ and $\mathbf{X_{a}^{i}}$ are multiplied to generate $\mathbf{X_{m}}$, which is passed it through the GWAP and MLP layers to generate the attention vector $\mathbf{f_{a}}$. Then, $\mathbf{f_{a}}$ and $\mathbf{X_{m}}$ are multiplied to extract boundary-guided global context information. Finally, BFM generates the boundary-fused feature $\mathbf{X_{f}^{i}}$ by summing boundary-guided global context information and $\mathbf{X_{a}^{i}}$ and applying $1 \times 1$ convolution.

\subsection{Objective Function}
Two types of supervision are applied: a camouflaged object mask $\mathbf{L_{co}}$ and a boundary map $\mathbf{L_{b}}$. First, weighted binary cross-entropy loss $\mathbf{L_{BCE}^w}$ and weighted IOU loss $\mathbf{L_{IOU}^w}$ are applied to $\mathbf{L_{co}}$, inspired by the works of~\cite{wei2020f3net, dong2021towards}, which helps assign more weight to the hard case pixels. In addition, binary cross-entropy loss $\mathbf{L_{BCE}}$ is applied to $\mathbf{L_{b}}$ and a $1 \times 1$ dilation kernel is employed for the ground truth boundary map to solve the lack of supervision signal due to the thin boundary map. Thus, the final objective function $\mathbf{L_{total}}$ is expressed as follows:

\begin{figure*}[t]
	\setlength{\belowcaptionskip}{-24pt}
	\begin{center}
		\includegraphics[width=0.90\linewidth]{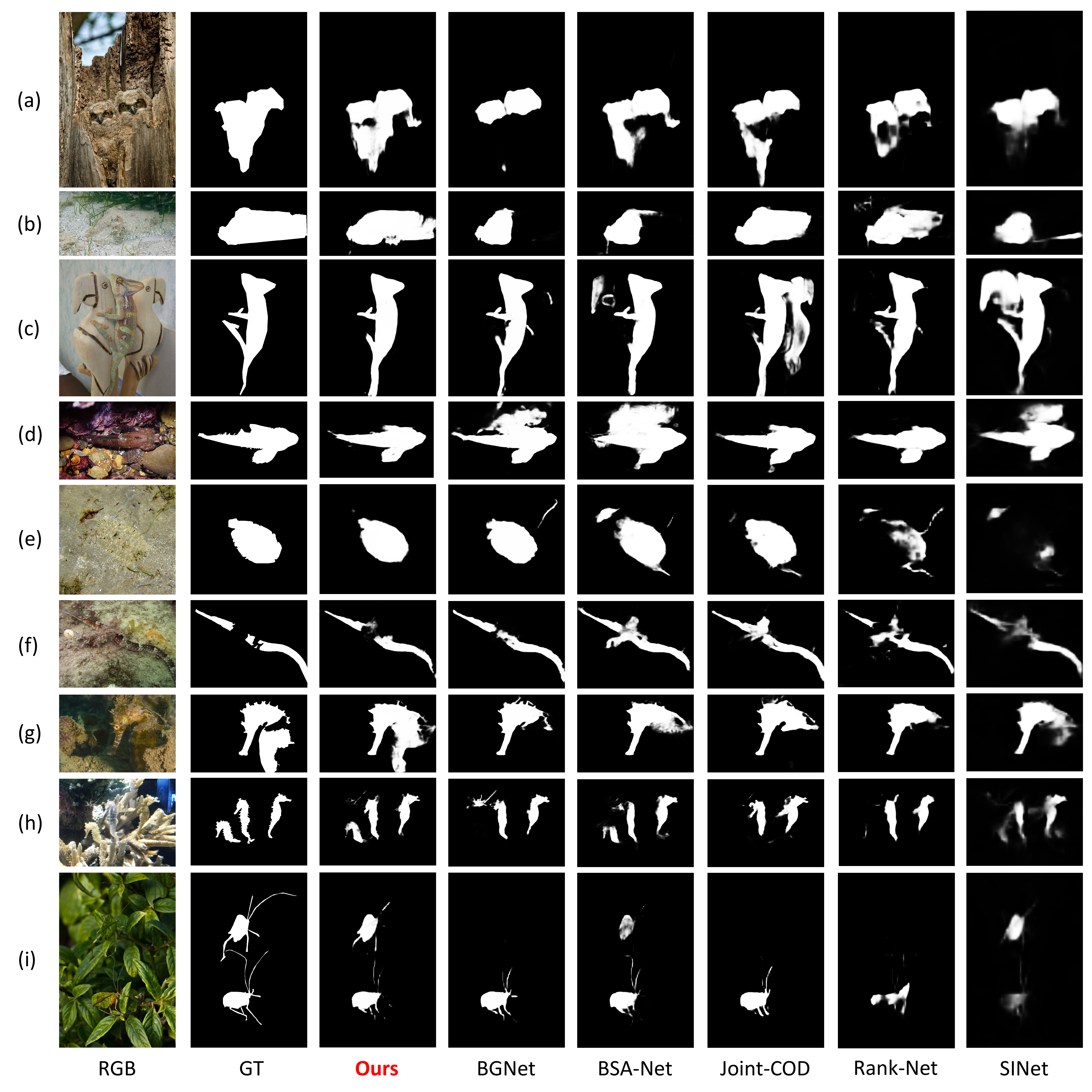}
		\caption{Qualitative comparison of the proposed method with previous state-of-the-art methods, BGNet~\cite{sun2022boundary}, BSA-Net~\cite{zhu2022can}, Joint-COD~\cite{li2021uncertainty}, Rank-Net~\cite{lv2021simultaneously}, and SINet~\cite{fan2020camouflaged}. The visualization result of our method is DPS-Net-R in Table~\ref{table:tb1}.}
		\vspace{-0.5cm}
		\label{fig:result}
	\end{center}
\end{figure*}

\begin{equation}
	\mathbf{L_{total}} = \mathbf{L_{co}}\left(\mathbf{I_{pred}}, \mathbf{I_{gt}}\right) + \mathbf{L_{b}}\left(\mathbf{E_{pred}}, \mathbf{E_{gt}}\right).
\end{equation}

\begin{table*}[t]
	\begin{center}
		\caption{Performance comparison of the proposed method with other state-of-the-art methods on the CAMO~\cite{le2019anabranch}, COD10K~\cite{fan2020camouflaged}, and CHAMELEON~\cite{skurowski2018animal} datasets. $\uparrow$ indicates that higher is better, and $\downarrow$ indicates that lower is better. \textbf{-R} stands for Res2Net50~\cite{gao2019res2net}, \textbf{-S} stands for Swin-S~\cite{liu2021swin}, \textbf{-P} stands for PVTv2-B4~\cite{wang2022pvt} backbone, pre-trained with ImageNet dataset~\cite{deng2009imagenet}.
		}
		\label{table:tb1}
		\resizebox{1.8\columnwidth}{!}{
			\begin{tabular}{lccccccccccccc}
				\hline
				\multirow{2}{*}{Method} & \multirow{2}{*}{Size} & \multicolumn{4}{c}{CAMO~\cite{le2019anabranch}} & \multicolumn{4}{c}{COD10K~\cite{fan2020camouflaged}}   & \multicolumn{4}{c}{CHAMELEON~\cite{skurowski2018animal}} \\ \cline{3-14} 
				& & $S_{\alpha}\uparrow$ & $E_{\xi}\uparrow$ & $F_{w}\uparrow$ & $M\,\downarrow$ & $S_{\alpha}\uparrow$ & $E_{\xi}\uparrow$ & $F_{w}\uparrow$ & $M\,\downarrow$ & $S_{\alpha}\uparrow$ & $E_{\xi}\uparrow$ & $F_{w}\uparrow$ & $M\,\downarrow$     \\ \hline
				CNN-Based Methods &&&&&&&&&&&&& \\ \hline
				SINet~\cite{fan2020camouflaged} & $352 \times 352$ & 0.751 &  0.771 & 0.606 & 0.100 & 0.771 & 0.806 & 0.551 &  0.051 & 0.869 & 0.891 & 0.740 & 0.044 \\
				TANet~\cite{ren2021deep} & $384 \times 384$ & 0.793 & 0.834 & 0.690 & 0.083 & 0.803 & 0.848 & 0.629 & 0.041 & 0.888 & 0.911 & 0.786 & 0.036 \\
				TINet~\cite{zhu2021inferring} & $352 \times 352$ & 0.781 & 0.847 & 0.678 & 0.087 & 0.793 & 0.848 & 0.635 & 0.043 & 0.874 & 0.916 & 0.783 & 0.038 \\
				C2F-Net~\cite{sun2021context} & $352 \times 352$ & 0.796 & 0.854 & 0.719 & 0.080 & 0.813 & 0.890 & 0.686 & 0.036 & 0.888 & 0.935 & 0.828 & 0.032 \\
				PFNet~\cite{mei2021camouflaged} & $416 \times 416$ & 0.782 & 0.852 & 0.695 & 0.085 & 0.800 & 0.868 & 0.660 & 0.040 & 0.882 & 0.942 & 0.810 & 0.033 \\
				R-MGL~\cite{zhai2021mutual} & - & 0.775 & 0.847 & 0.673 & 0.088 & 0.814 & 0.865 & 0.666 & 0.035 & 0.893 & 0.923 & 0.813 & 0.030 \\
				Rank-Net~\cite{lv2021simultaneously} & $352 \times 352$ & 0.708 & 0.755 & 0.645 & 0.105 & 0.760 & 0.831 & 0.658 & 0.045 & 0.842 & 0.896 & 0.794 & 0.046 \\
				Joint-COD~\cite{li2021uncertainty} & $352 \times 352$ & 0.803 & 0.853 & - & 0.076 & 0.817 & 0.892 & - & 0.035 & 0.894 & 0.943 & - & 0.030 \\
				UGTR~\cite{yang2021uncertainty} & -  & 0.785 & 0.859 & 0.686 & 0.086 & 0.818 & 0.850 & 0.667 & 0.035 & 0.888 & 0.918 & 0.796 & 0.031 \\
				ERRNet~\cite{ji2022fast} & $352 \times 352$ & 0.761 & 0.817 & 0.660 & 0.088 & 0.780 & 0.867 & 0.629 & 0.044 & 0.877 & 0.927 & 0.805 & 0.036 \\
				CANet~\cite{liu2022modeling} & $480 \times 480$ & 0.807 & 0.866 & 0.767 & 0.075 & 0.832 & 0.890 & 0.745 & 0.032 & 0.901 & 0.940 & 0.843 & 0.028 \\
				BSA-Net~\cite{zhu2022can} & $384 \times 384$ & 0.796 & 0.851 & 0.717 & 0.079 & 0.818 & 0.891 & 0.699 & 0.034 & 0.895 & 0.946 & 0.841 & 0.027 \\
				SegMaR~\cite{jia2022segment} & $352 \times 352$ & 0.815 & 0.872 & 0.742 & 0.071 & 0.833 & 0.895 & 0.724 & 0.033 & 0.906 & 0.954 & 0.860 & 0.025 \\
				BGNet~\cite{sun2022boundary} & $416 \times 416$ & 0.812 & 0.870 & 0.749 & 0.073 & 0.831 & 0.901 & 0.722 & 0.033 & -     & -     & -     & -     \\
				\textbf{DPS-Net-R (Ours)} & $352 \times 352$ & 0.818 & 0.885 & 0.776 & 0.071 & 0.836 & 0.904 & 0.752 & 0.031 & 0.910 & \underline{0.960} & 0.878 & 0.023 \\ \hline
				Transformer-Based Methods &&&&&&&&&&&&& \\ \hline
				COS-T~\cite{wang2022camouflaged} & $512 \times 512$ & 0.813 & 0.896 & 0.776 & 0.060 & 0.790 & 0.901 & 0.693 & 0.035 & 0.885 & 0.948 & 0.854 & 0.025 \\
				TPRNet~\cite{zhang2022tprnet} & $352 \times 352$ & 0.814 & 0.870 & 0.781 & 0.076 & 0.829 & 0.892 & 0.725 & 0.034 & 0.891 & 0.930 & 0.816 & 0.031 \\
				DTINet~\cite{liu2022boosting} & - & 0.857 & 0.912 & 0.796 & 0.050 & 0.824 & 0.893 & 0.695 & 0.034 & 0.883 & 0.928 & 0.813 & 0.033 \\
				\textbf{DPS-Net-S (Ours)} & $352 \times 352$ & \underline{0.871}  & \textbf{0.933} & \underline{0.835} & \underline{0.048} & \underline{0.859} & \underline{0.933} & \underline{0.782} & \underline{0.024} & 0.905 & \underline{0.960} & \underline{0.879} & \underline{0.023} \\
				\textbf{DPS-Net-P (Ours)} & $352 \times 352$ & \textbf{0.878}  & \underline{0.931} & \textbf{0.840} & \textbf{0.044} & \textbf{0.875} & \textbf{0.936} & \textbf{0.797} & \textbf{0.022} & \textbf{0.919} & \textbf{0.961} & \textbf{0.880} & \textbf{0.021} \\ \hline
			\end{tabular}
		}
	\end{center}
\vspace{-0.5cm}
\end{table*}

\section{Experiments}
\subsection{Datasets}
We perform experiments on three popular COD benchmarks to validate the effectiveness of the proposed method: CHAMELEON~\cite{skurowski2018animal}, CAMO~\cite{le2019anabranch}, and COD10K~\cite{fan2020camouflaged}. CHAMELEON~\cite{skurowski2018animal} is a small dataset containing only 76 images, which are collected from the Internet. The CAMO~\cite{le2019anabranch} dataset includes 1250 images (1000 images in the train set and 250 images in the test set). Finally, COD10K~\cite{fan2020camouflaged} is the largest dataset, containing 10000 images with 10 super-classes and 78 sub-classes collected from websites. Following the method of~\cite{fan2020camouflaged, sun2022boundary, lv2021simultaneously, zhu2022can, mei2021camouflaged}, we use images with camouflaged objects in the experiments, in which 3040 images from COD10K~\cite{fan2020camouflaged} and 1000 images from CAMO~\cite{le2019anabranch} are used for training, and the remaining images are employed for testing.

\subsection{Evaluation Metrics}
We evaluate the performance of our method by employing four evaluation metrics: mean absolute error (MAE, $M$), S-measure ($S_{\alpha}$)~\cite{fan2017structure}, E-measure ($E_{\xi}$)~\cite{fan2018enhanced}, and weight F-measure ($F_{w}$)~\cite{achanta2009frequency}.

\subsection{Implementation Details}
The patch size of the proposed model and the number of reference points are set to $12 \times 12$ and $3 \times 3$, respectively. We implement the proposed method using the open deep-learning framework PyTorch. All the input images are resized to $352 \times 352$ and augmented by randomly horizontal flipping. During the training stage, we used the Adam optimizer~\cite{kingma2014adam} with $w_1=0.9$, $w_2=0.999$, and $\epsilon =10 ^ { -8 }$. The learning rate decayed from $10^{-4}$ to $10^{-5}$ with the cosine annealing scheduler~\cite{loshchilov2016sgdr}. Further, we set the total number of epochs to 200 with a batch size 16. Two NVIDIA RTX 3090 GPUs are used for all experiments in this study.

\subsection{Results}

\noindent
\textbf{Quantitative Results.} Table~\ref{table:tb1} shows the quantitative results of the proposed DPS-Net. The proposed model is evaluated with Res2Net50~\cite{gao2019res2net}, Swin-S~\cite{liu2021swin}, and PVTv2-B4~\cite{wang2022pvt} backbone encoders. To ensure a fair comparison, we compare the models separately based on the encoder architecture adopted, either CNN or transformer. As shown in Table~\ref{table:tb1}, the proposed model achieves state-of-the-art performance on all datasets and exhibits significant performance improvement, especially when transformer encoder is adopted. Furthermore, compared to BSA-Net~\cite{zhu2022can} and BGNet~\cite{sun2022boundary}, which use similar boundary reconstruction-based methods and the same Res2Net50~\cite{gao2019res2net} encoder, DPS-Net-R obtains higher performance despite the smaller test image resolution. This shows that the proposed method can extract information about the disguised object more effectively than the existing boundary-reconstruction method. We demonstrate the effectiveness of the proposed model through various ablation studies in Section~\ref{ablation}.

\noindent
\textbf{Qualitative Results.} In Figure~\ref{fig:result}, we present a comparison of the qualitative results of DPS-Net-R with five state-of-the-art COD approaches, BGNet~\cite{sun2022boundary}, BSA-Net~\cite{zhu2022can}, Joint-COD~\cite{li2021uncertainty}, Rank-Net~\cite{lv2021simultaneously}, and SINet~\cite{fan2020camouflaged}, in various challenging scenarios, including multiple objects, low contrast, thin objects, and long distance. As depicted in Figure~\ref{fig:result} (b) and (e), when the target object's texture is very similar to that of the background, our proposed model produces an accurate prediction map. Additionally, the proposed method demonstrates robustness in scenes with multiple camouflaged objects, as shown in (h) and (i). This is because the DPS transformer of the proposed model learns the relationship between patches to enhance long-distance connectivity, and the aggregator efficiently integrates them. Furthermore, even when thin and long objects are present, as in (f) and (i), our deformable point sampling method can extract useful information, such as edges, for generating prediction maps. We will discuss the usefulness of the deformable point sampling methods in detail in Section~\ref{ablation}.

\begin{table*}[t]
	\begin{center}
		\caption{Performance with different combinations of our contributions on the CAMO~\cite{le2019anabranch}, COD10K~\cite{fan2020camouflaged}, and CHAMELEON~\cite{skurowski2018animal} datasets. $\uparrow$ indicates that higher is better, and $\downarrow$ indicates that lower is better.
		}
		\label{table:tb2}
		\resizebox{2\columnwidth}{!}{
			\begin{tabular}{c|ccccc|cccc|cccc|cccc}
				\hline
				\multirow{2}{*}{Index} & \multicolumn{5}{c|}{Component} & \multicolumn{4}{c|}{CAMO~\cite{le2019anabranch}} & \multicolumn{4}{c|}{COD10K~\cite{fan2020camouflaged}} & \multicolumn{4}{c}{CHAMELEON~\cite{skurowski2018animal}} \\ \cline{2-18} 
				& Baseline & MFFM & DPS Transformer & Boundary Decoder & BFM & $S_{\alpha}\uparrow$ & $E_{\xi}\uparrow$ & $F_{w}\uparrow$ & $M\,\downarrow$ & $S_{\alpha}\uparrow$ & $E_{\xi}\uparrow$ & $F_{w}\uparrow$ & $M\,\downarrow$ & $S_{\alpha}\uparrow$ & $E_{\xi}\uparrow$ & $F_{w}\uparrow$ & $M\,\downarrow$     \\ \hline
				(a)                    & \ding{51}      &      &                 &                  &     & 0.791 & 0.863 & 0.746 & 0.084 & 0.800 & 0.889 & 0.737 & 0.035 & 0.867 & 0.940 & 0.839 & 0.033 \\
				(b)                    & \ding{51}      & \ding{51}  &                 &                  &     & 0.792 & 0.865 & 0.748 & 0.083 & 0.800 & 0.890 & 0.734 & 0.035 & 0.870 & 0.942 & 0.840 & 0.031 \\
				(c)                    & \ding{51}      & \ding{51}  &                 & \ding{51}              &     & 0.794 & 0.862 & 0.754 & 0.082 & 0.800 & 0.885 & 0.734 & 0.033 & 0.875 & 0.943 & 0.840 & 0.031 \\
				(d)                    & \ding{51}      & \ding{51}  &                 & \ding{51}              & \ding{51} & 0.801 & 0.868 & 0.757 & 0.080 & 0.809 & 0.894 & 0.735 & 0.034 & 0.879 & 0.949 & 0.841 & 0.032 \\
				(e)                    & \ding{51}      & \ding{51}  & \ding{51}             &                  &     & 0.805 & 0.873 & 0.760 & 0.079 & 0.817 & 0.892 & 0.733 & 0.033 & 0.885 & 0.946 & 0.850 & 0.030 \\
				(f)                    & \ding{51}      & \ding{51}  & \ding{51}             & \ding{51}              &     & 0.812 & 0.883 & 0.766 & 0.073 & 0.823 & 0.895 & 0.748 & 0.032 & 0.894 & 0.955 & 0.866 & 0.026 \\
				(g)                    & \ding{51}      & \ding{51}  & \ding{51}             & \ding{51}              & \ding{51} & \textbf{0.818} & \textbf{0.885} & \textbf{0.776} & \textbf{0.071} & \textbf{0.836} & \textbf{0.904} & \textbf{0.752} & \textbf{0.031} & \textbf{0.910} & \textbf{0.960} & \textbf{0.878} & \textbf{0.023} \\ \hline
			\end{tabular}
		}
	\end{center}
\vspace{-0.5cm}
\end{table*}

\subsection{Ablation Analysis}
\label{ablation}

We verify the performance of our model through various ablation studies. All experiments are conducted with DPS-Net-R. Table~\ref{table:tb2} presents the effects of the proposed modules in various combinations. In particular, the baseline refers to a model comprising a simple encoder and decoder.

\noindent
\textbf{Effect of MFFM.} As revealed by (a) and (b) in Table~\ref{table:tb2}, the use of the proposed MFFM improves the performance compared to the baseline model. This is because the parallel dilated convolution integrates multi-scale contextual features and increases the size of the receptive field.

\begin{figure}[t]
	\setlength{\belowcaptionskip}{-24pt}
	\begin{center}
		\includegraphics[width=1.0\linewidth]{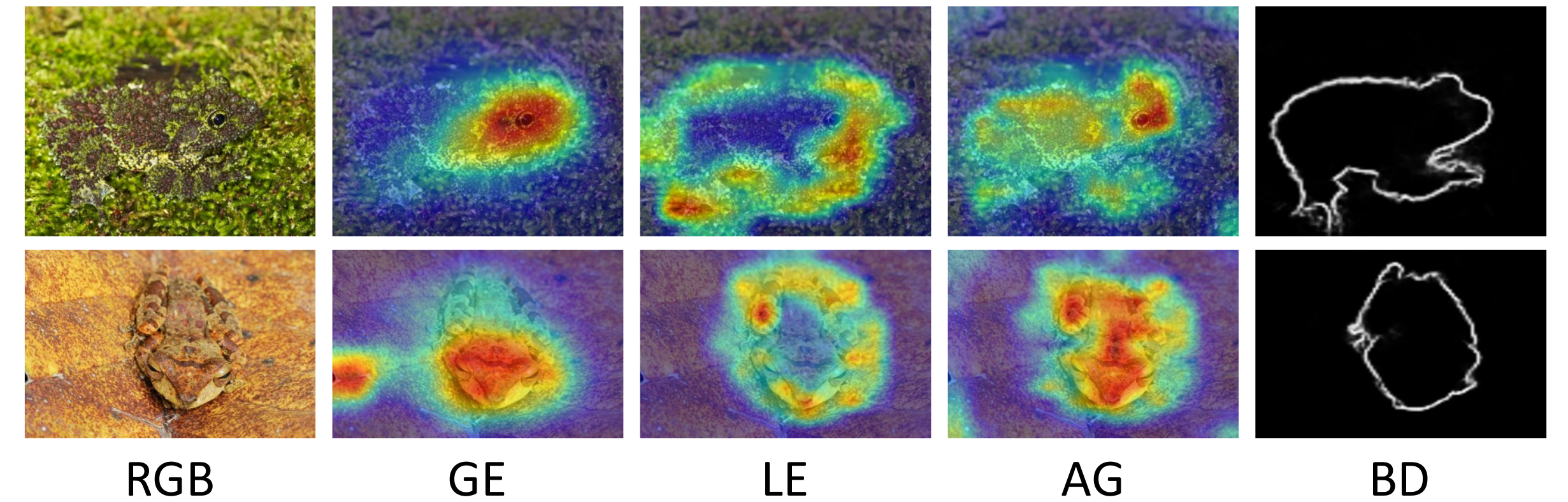}
		\caption{Visualization of the input image (RGB) and features extracted using a global extractor (GE), features extracted using a local extractor based on deformable point sampling (LE), features integrated using an aggregator (AG), and the predicted boundary map (BD).}
		\vspace{-0.5cm}
		\label{fig:edge}
	\end{center}
\end{figure}

\noindent
\textbf{Effect of DPS Transformer.} The (d) and (g) of Table~\ref{table:tb2} demonstrate the effectiveness of the proposed DPS transformer, which shows that DPS transformer can significantly enhance the COD performance. The DPS transformer is located between the encoder and boundary decoder, supplementing the encoder's insufficient feature extraction ability and refining the features before passing them to the boundary decoder, which helps improve performance. Figure~\ref{fig:edge} visualizes the effects of each part of the DPS transformer. As shown in the figure, the global extractor focuses on rough global spatial information based on salient keypoints of the target object. On the other hand, the local extractor focuses on local information of boundaries to distinguish between background and objects based on deformable sampling. Since the information extracted by the two extractors is clearly different, the proposed aggregator effectively fuses these two types of information to generate features that are unique to disguised objects that are distinguished from the background. As shown in Figure~\ref{fig:intro}, this complements the feature extraction capability of the encoder, and enables the model to generate accurate boundary maps.

\begin{figure}[t]
	\setlength{\belowcaptionskip}{-24pt}
	\begin{center}
		\includegraphics[width=0.95\linewidth]{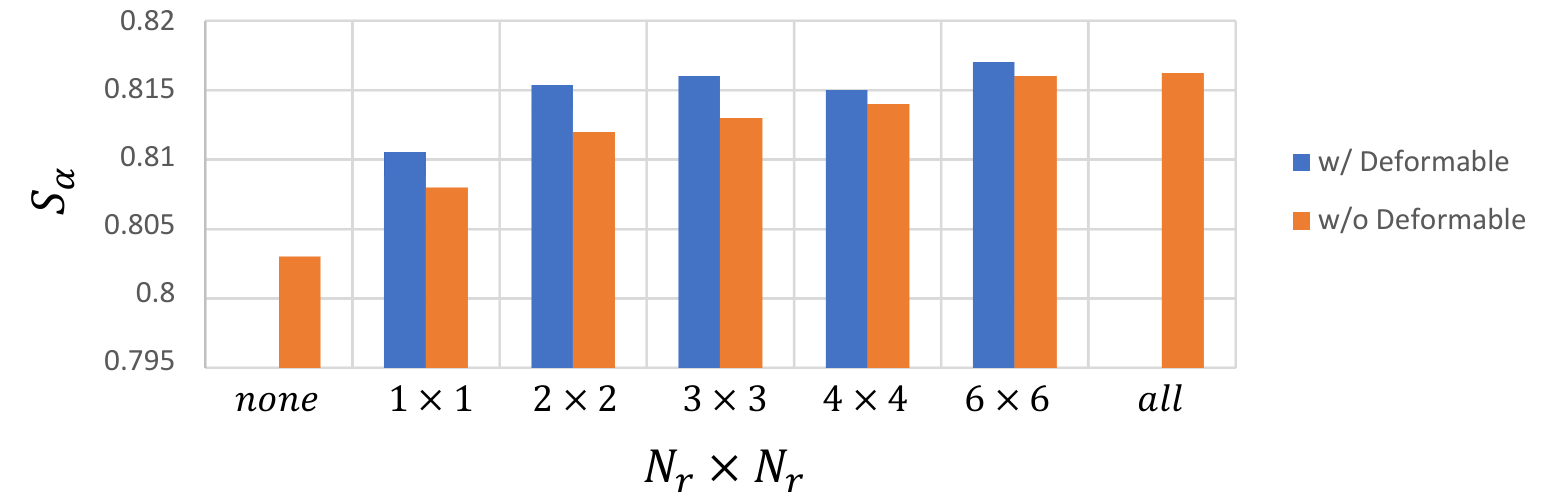}
		\caption{Comparison of S-measure of models on the CAMO~\cite{le2019anabranch} dataset according to the number of reference points $N _ { r } \times N _ { r }$ and whether deformable sampling is applied in the DPS transformer. Here, ``none'' indicates that the DPS transformer is not used ((d) in Table~\ref{table:tb2}) and ``all'' indicates that all pixels should be used in the patch as reference points.}
		\label{fig:ref_num}
	\end{center}
\end{figure}

\noindent
\textbf{Effect of Deformable Point Sampling Method.} Figure~\ref{fig:ref_num} describes the results of additional experiments on the deformable point sampling method of the local extractor of the DPS transformer. The number of sampled reference points is changed to $0 \times 0$, $1 \times 1$, $2 \times 2$, $3 \times 3$, $4 \times 4$, $6 \times 6$. Additionally we describe the result of sampling all pixels. As shown in Figure~\ref{fig:ref_num}, when using the deformable point sampling method, the performance generally increases as the number of sampled pixels increases, but does not show a significant increase beyond $3 \times 3$. Furthermore, COD performance is improved by using the deformable method without fixing the reference point. This shows that the proposed method, which limits the number of sampling points and affords sampling position freedom, reduces the computational amount and is effective. When using fixed reference points, the performance increases as the number of points increases; however, this process is inefficient because it requires considerably more points than the deformable method does.

\noindent
\textbf{Effect of Boundary Decoder and BFM.} In Table~\ref{table:tb2}, (e), (f), and (g) present the effects of the proposed boundary decoder and BFM. If only the boundary decoder is used without BFM, the generated boundary map is concatenated with the decoder features without BFM. The results show that, using the boundary map generated by the boundary decoder as a guide, boundary context information is extracted and effectively fused by BFM. This shows that boundary prediction map generation aids in the model's ability to distinguish between the delicate foreground and background boundaries of camouflaged objects.

\noindent
\textbf{Compare model efficiency with other methods.} As shown in Table~\ref{table:tb1}, the proposed DPS-Net-R achieves state-of-the-art performance in methods based on CNN encoders, but shows performance similar to SegMaR~\cite{jia2022segment} in some evaluation metrics. However SegMaR~\cite{jia2022segment} employ iterative multi-stage structure, i.e., the predicted masks are used for mask prediction again. This structure can generate accurate prediction masks by refining the prediction mask step by step, but it has a significant drawback of requiring heavy computational burden. The Table~\ref{table:FPS} compares the inference time (FPS) and number of parameters of SegMaR~\cite{jia2022segment}, BGNet~\cite{sun2022boundary}, and ours. The proposed DPS-Net-R has the least number of parameters and the fastest inference speed. Furthermore, DPS-Net-P based on transformer encoder shows similar inference speed to SegMaR~\cite{jia2022segment}, but exhibits significantly higher mask prediction performance. The quantitative comparison demonstrates the efficiency of our proposed approach.

\begin{table}[t]
	\begin{center}
		\caption{Comparison of the number of parameters and FPS with state-of-the-art models.}
		\label{table:FPS}
		\resizebox{1\columnwidth}{!}{
			\begin{tabular}{c|cccc}
				\hline
				& SegMaR~\cite{jia2022segment} & BGNet~\cite{sun2022boundary} & DPS-Net-R & DPS-Net-P \\ \hline
				Backbone	   & ResNet50        & Res2Net50       & Res2Net50 & PVTv2-B4\\
				FPS            & 13.5        & 31.3       & \textbf{36.3} & 12.7\\
				Parameters (M) & 56.2        & 79.9       & \textbf{41.9} & 76.7\\ \hline
			\end{tabular}
		}
	\end{center}
	\vspace{-0.5cm}
\end{table}

\section{Conclusion}
We propose the DPS-Net for camouflaged object detection. Our method utilizes a DPS transformer to directly capture sparse supervision signals for object boundary areas from the encoder features. The DPS transformer effectively complements the feature extraction capability of the encoder and integrates global information and local boundary information, allowing for accurate prediction maps. The proposed method demonstrates state-of-the-art performance on three datasets. Our method has the potential for various applications, and future work may involve exploring the DPS transformer's potential in other domains.

{\small
\bibliographystyle{ieee_fullname}
\bibliography{egbib}
}

\end{document}